\documentclass[preprint,12pt]{elsarticle}

\usepackage[dvipsnames]{xcolor}
\usepackage{amssymb}
\usepackage[section]{placeins}
\usepackage{amsmath}
\usepackage[hyphens]{url}
\usepackage[hidelinks]{hyperref}
\hypersetup{breaklinks=true}
\usepackage[onehalfspacing]{setspace}
\usepackage{multicol}
\usepackage{verbatim}

\usepackage{graphicx}
\usepackage[hyphens]{url}
\usepackage{listings}
\usepackage{color}

\definecolor{dkgreen}{rgb}{0,0.6,0}
\definecolor{gray}{rgb}{0.5,0.5,0.5}
\definecolor{mauve}{rgb}{0.58,0,0.82}

\journal{XX}

\begin{document}

\begin{frontmatter}

\title{Exploring Flip Flop memories and beyond: training recurrent neural networks with key insights}

%\maketitle
\author[label1,label2,label3]{Cecilia Jarne}
\affiliation[label1]{organization={Departamento de Ciencia y Tecnologia de la Universidad Nacional de Quilmes},%Department and Organization
            %addressline={}, 
            city={Bernal},
            %postcode={}, 
            state={Buenos Aires},
            country={Argentina}}
            
\affiliation[label2]{organization={CONICET},
            %city={Bernal},
            %postcode={}, 
            state={Buenos Aires},
            country={Argentina}}
            
\affiliation[label3]{organization={Center of Functionally Integrative Neuroscience, Department of Clinical Medicine, Aarhus University},
            %city={Bernal},
            %postcode={}, 
            state={Aarhus},
            country={Denmark}}   
\begin{abstract}

Training neural networks to perform different tasks is relevant across various disciplines. In particular, Recurrent Neural Networks (RNNs) are of great interest in Computational Neuroscience. Open-source frameworks dedicated to Machine Learning, such as Tensorflow \cite{chollet2015keras} and Keras \cite{tensorflow2015-whitepaper} have produced significant changes in the development of technologies that we currently use. This work aims to make a significant contribution by comprehensively investigating a temporal processing task, specifically a 3-bit Flip Flop memory. We delve into the entire modelling process, encompassing equations, task parametrization, and software development. The obtained networks are meticulously analyzed to elucidate dynamics, aided by an array of visualization and analysis tools. Moreover, the provided code is versatile enough to facilitate the modeling of diverse tasks and systems. Furthermore, we present how memory states can be efficiently stored in the vertices of a cube in the dimensionally reduced space, supplementing previous results with a distinct approach.

\end{abstract}

\end{frontmatter}

\section{Introduction} \label{intro}

\Urlmuskip=0mu plus 1mu\relax

Machine learning methods, and in particular Deep Learning, have demonstrated remarkable success in a wide range of tasks in multiple domains \cite{Ogunfunmi2019}. In recent years, the emergence of open-source frameworks dedicated to Machine Learning, such as Pytorch, Tensorflow and Keras \cite{NEURIPS2019_9015,chollet2015keras, tensorflow2015-whitepaper} has produced huge changes in the development of technologies that we use every day. {Due to their novelty and complexity, it can take time to properly utilize these frameworks in other relevant scientific domains, such as Computational Neuroscience or fields related to engineering.}

{This enforces the need to develop clear tutorials or premiers to explain how to implement the algorithms to address the scientific questions of interest and how to solve different problems using the newly available tools.}

One relevant problem is how to build models for the study of dynamical systems and how to extract relevant information.

Neural Networks are algorithms that allow us to model different systems. According to the Universal Approximation Theorem, a neural network with one hidden layer containing a sufficient but finite number of neurons can approximate any continuous function to a reasonable accuracy under certain conditions for activation functions \cite{HORNIK1991251}. {This theorem has been extended to RNNs. It is well known that} dynamical system can be approximated by continuous-time RNNs \cite{FUNAHASHI1993801}.

The problem of training neural networks to perform different tasks is relevant across various disciplines that go beyond Machine Learning. In particular, RNNs are of great interest in different scientific communities. They are widely used in {Computational} Neurosciences to describe the behaviour of the cortex, an area that presents great recurrence in its connections \cite{Murphy2009}. These models also have great relevance concerning control systems and other areas such as electronics \cite{DENG2013281, 7966138, DINH201444}. {Particularly those networks are related to the processing of temporal information and the production of time-dependent outputs.}

{In the field of Machine learning, more sophisticated architectures such as LSTM (Long Short Term Memory units) or GRU (Gated recurrent units) are widely spread and have been used to process temporal sequences since they do not have the same limitations as RNNs to process long time dependencies \cite{69e088c8129341ac89810907fe6b1bfe, NIPS2015_5955, pascanu2013, 279181, 10.1162/neco_a_01253}.}

{However, a simple RNN model still constitutes a vast field of study. The primary reason for this is that it is employed in Computational Neuroscience to comprehend computation in terms of collective dynamics, which is involved in cognitive tasks like motor control, temporal brain processing tasks, decision-making, or working memory \cite{doi:10.1146/annurev-neuro-092619-094115, Jarne2023_2}. These models are important because they are  {useful for describing} the behaviour of brain areas, such as the prefrontal cortex.}

RNNs also allow the incorporation of realistic characteristics at the biological level, such as Dale's law \cite{doi:10.1177/003591573502800330, Jarne2023_2}, sparsity or different characteristics of interest in animal models.

There are general tutorials available on artificial neural networks, such as \cite{YANG20201048}. However, in this work, we will focus extensively on recurrent neural networks and their application in computational neuroscience. Recurrent neural networks play a relevant role in understanding complex neural processes and dynamics. Throughout this tutorial, we will delve into the architecture, training methodologies, and practical implementation of recurrent neural networks, exploring their significance and potential contributions to the field of computational neuroscience.

In this work, the modelling of these complex systems is presented using Tensorflow and Keras. The reason for this selection is that such scientific libraries are open-source, their use is rapidly growing, {and they become increasingly popular. One can find excellent documentation for software development about them} \cite{gulli2017deep, 10.5555/3235300, singh2019learn}, and also we have new tools such as Google Colaboratory that allows implementing and testing models directly online.

A simple RNN was chosen because it is relevant in various fields. Here it was trained to perform a time-series processing task inspired in Computational Neuroscience studies \cite{SUSSILLO2014156}. The implementation of the network, the training, and the tools are carefully described here, as well as different forms to obtain the information that allows a suitable description of the system under study.

Training a RNN to perform temporal tasks has many difficulties and can be done through various paradigms. Here it is proposed to approach the problem through supervised learning. The entire procedure is described in detail.

{The Flip Flop task was chosen as a case example. On one hand, a Flip Flop is the simplest sequential system that one can build \cite{floyd2003digital}. In fact, a 3-bit memory was studied, which is a task composed of a set of Flip Flops. This is also a working memory task considered previously in other works in the field of Computational Neuroscience \cite{DBLP:journals/neco/SussilloB13, BARAK20171, Jarne2022}.} The parameterization of the task is as described in \cite{DBLP:journals/neco/SussilloB13}, but it is also revisited here. Gradient descendant minimization was used to take advantage of different optimized implementations of the current algorithms available. 

{Every step is thoroughly explained, from parameterizing the task to describing the dynamics of trained networks.} This example is used to show how the problem of training networks can be studied using these computing tools applied in any temporal task in general, but also to discuss the limitations that networks have and the alternatives to solve them.

{The rest of the paper is organized as follows. In Section \ref{model}, the description of the dynamics, discretization and code examples are presented. In Section \ref{task}, the task parametrization is shown. Section \ref{traning} describes the training protocol. In Section \ref{results}, the results, different analyses of the network, tools and software are discussed in detail. Finally, Section \ref{conclu} includes the final remarks.}

\section{Model} \label{model}

\Urlmuskip=0mu plus 1mu\relax

The dynamics of the units in the RNN model is inspired by equation \ref{eq-01} \cite{Hopfield3088}, where units have index $i$, with $i=1,2...,n$.

\begin{center}
\begin{equation}
\frac{dh_i(t)}{dt}=-\frac{h_i(t)}{\tau}+\sigma \left( \sum_{j}w^{Rec}_{ij}h_j(t)+\sum_{j}w^{in}_{ij} x_j(t) \right)
\label{eq-01}
\end{equation} 
\end{center}

In Equation \ref{eq-01}, the matrix elements $w^{Rec}_{ij}$ are the synaptic connection strengths of the matrix $\mathbf{W^{Rec}}$ and $w^{in}_{ij}$ the matrix elements of $\mathbf{W^{in}}$ from the input units. $x_j(t)$ are the components of the vector $\mathbf{X}(t)$ of the input signal. $\tau$ represents the time constant of the system and $\sigma$ is a non-linear activation function. 

The network is fully connected, and matrices have weights given by a certain parametrization of interest. In this case, {we considered} a normal distribution with zero mean and variance $\frac{1}{N}$.

The network has three layers:  the input, the recurrent hidden layer, and the output layer. The readout, in terms of the matrix elements $w^{out}_{ij}$, from $\mathbf{W^{out}}$ is described by Equation \ref{eq-02}.

\begin{center}
\begin{equation}
\mathbf{Z(t)}= \sum_{j}w^{out}_{ij}h_j(t)
\end{equation}
\label{eq-02}
\end{center}

\begin{center}
\begin{figure}[htb!]
%\hspace{1cm}\includegraphics[width=10cm]{figs/02_training.png}
\hspace{1cm}\includegraphics[width=11.5cm]{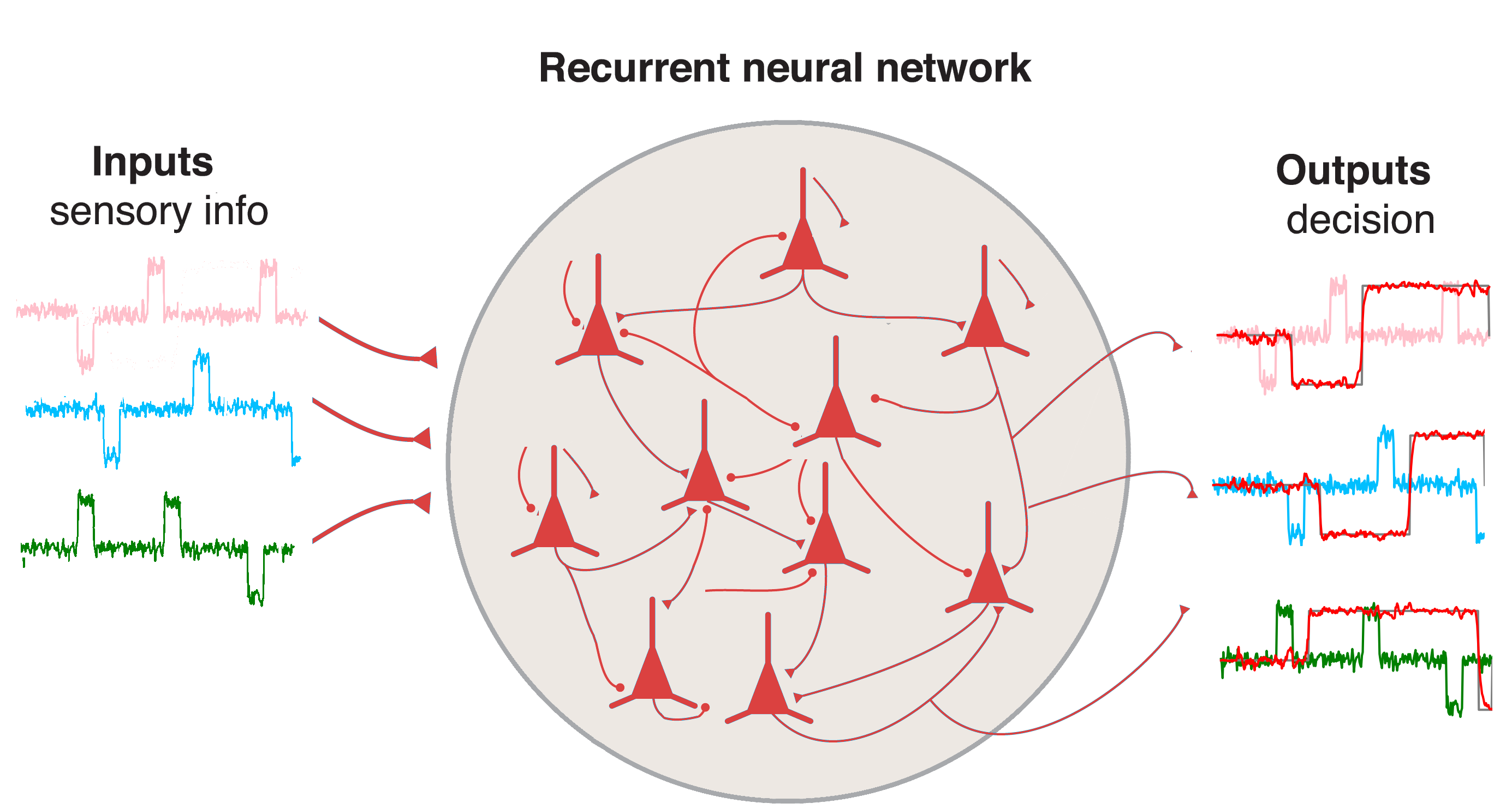}
\vspace{0.5cm}
\caption{{RNN schema that represents the network described by Equations \ref{eq-01} and \ref{eq-02}. In this case, we have three inputs and outputs to build the memory states for the 3-bit Flip Flop task.}}
\label{fig:1}  
\end{figure}
\end{center}

For this work, it was considered $\sigma () = tanh () $ and $ \tau = 1 $ without loss of generality. The model is discretized using Euler's method {following \cite{,Jarne2022, 10.1371/journal.pcbi.1007655, 10.1371/journal.pone.0220547,Bi10530}}. A simple schema of the model is presented in Figure \ref{fig:1}. In this case, {the network have} three inputs and three outputs corresponding to the inputs and memory states of the 3-bit Flip Flop task.

As described in Section \ref{intro}, the model is implemented in Python using Keras and Tensorflow \cite{chollet2015keras,tensorflow2015-whitepaper}. This allows {us} to use all current algorithms and optimization {methods} developed and maintained by a massive research community.

In vector form, the equations \ref{eq-01} and \ref{eq-02} can be written as:

\begin{center}
\begin{equation}
\frac{d\mathbf{H}(t)}{dt}=-\frac{\mathbf{H}(t)}{\tau}+\sigma(\mathbf{W^{Rec}}\mathbf{H}(t)+\mathbf{W^{in}}\mathbf{X}(t))
 \label{eq-01b}
\end{equation}
\end{center}

and respectively:

\begin{center}
\begin{equation}
\mathbf{Z(t)}=\mathbf{ W^{out}}\mathbf{H}(t)
\label{eq-02b}
\end{equation}
\end{center}

The system represented by Equation \ref{eq-01} is approximated using Euler's method, as previously indicated, with a step time $\delta t$. A value of $\tau=1$ was considered. Then the dynamics of the discrete-time RNN is giben by Equation \ref{eq-03}.

\begin{center}
\begin{equation}
\mathbf{H}(t+\delta t)=\mathbf{H}(t)+(-\mathbf{H}(t)+\sigma(\mathbf{W^{Rec}}\mathbf{H}(t)+\mathbf{W^{in}}\mathbf{X}(t))),
\label{eq-03}
\end{equation}
\end{center}

The {value considered for the time step is $\delta t=1$  to obtain the time evolution}. {Usually, the amplitude of the activity $\mathbf{H}(t)$ is adimensional or expressed in arbitrary units. It will depend on context}. Then, from Equation \ref{eq-03}, the activity of the recurrent units at the next time step is given by Equation \ref{eq-06}.

\begin{center}
\begin{equation}
\mathbf{H}(t+1)=\sigma(\mathbf{W^{Rec}}\mathbf{H}(t)+\mathbf{W^{in}}\mathbf{X}(t)))
\label{eq-06}
\end{equation}
\end{center}

Tensorflow has a recurrent layer directly implemented to represent Equation \ref{eq-06}, where it is possible to choose the initialization of the parameters,  {number of} units and activation function. This is shown in the following code box.

\begin{lstlisting}
tf.keras.layers.SimpleRNN( units,  activation="tanh",   kernel_initializer="glorot_uniform", recurrent_initializer="orthogonal",**kwargs)
\end{lstlisting}
\begin{small}
Code for a Recurrent layer defined in Tensor Flow.\label{code-1}\\
\end{small} 

{The time scale of the Equation \ref{eq-06} is arbitrary. If we are interested in scales related to cognitive processes, we can consider, for example, 1 ms of temporal resolution.}

{The RNN can be initialized with different weight distributions. Several options can be selected in TensorFlow. This choice will depend, on the one hand, on the existence of some physical motivation or hypothesis of the models. On the other hand, from the ML side, it will depend on the performance associated with the considered initialization.}

To define a Recurrent Network with the topology {shown} in Figure \ref{fig:1}, it is possible to build a sequential  model with layers such as:

\begin{lstlisting}
model = Sequential() 
model.add(SimpleRNN(units=N, input_shape=(None,3),activation="tanh"))
model.add(Dense(units=3,input_dim=N))
\end{lstlisting}
\begin{small}
Code for the sequential model defined for the network in Figure \ref{fig:1}.\label{code-2}\\
\end{small}

Where {$\texttt{input\_shape=(None,3)}$ means the shape of the input vector, $\texttt{activation='tanh'}$ corresponds to the definition of activation function, and $\texttt{Dense}$ is a fully connected output layer. In this way, we completed the first step which is of the model definition in terms of the code.

{Other network architectures, such as Gated Recurrent Units \cite{69e088c8129341ac89810907fe6b1bfe} or Long Short Term Memories \cite{NIPS2015_5955}, could be selected if there was any motivation from the perspective of the mechanisms to take into account. Both are already implemented in TensorFlow. Such code options are shown in the boxes below.}

\vspace{0.5cm}
\begin{lstlisting}
model = Sequential() 
model.add(layers.GRU(units=N, return_sequences=True))
...
\end{lstlisting}
\begin{small}
Code using other architecture (GRU) for the sequential model defined for Figure \ref{fig:1}.\label{code-2b}\\
\end{small}

\begin{lstlisting}
model = Sequential() 
model.add(layers.LSTM(units=N, input_shape=(None,3)))
...
\end{lstlisting}
\begin{small}
Code using other architecture (LSTM) for the sequential model defined for Figure \ref{fig:1}.\label{code-2c}\\
\end{small}

{The choice of the appropriate architecture will depend on the system to be modelled. Particular features, such as bias terms, can be also considered. In some cases it is possible, using Keras and Tensorflow as development tools, to build architectures with additional features that are not pre-defined. This can be done by using the class structures in the framework.}

{\section{Task selection and parametrization} \label{task}}

The parameterization of the task to be studied will have strong consequences on the possible dynamics obtained from the system through network training.

{Previous works have considered some relevant tasks in Computational Neuroscience related to decision-making or working memory. For example in \cite{jarne2019detailed, 10.1088/2632-072X/abdee3, Jarne2022}. All these processes use time-varying signals, which are very different from the binary boolean operations considered with forward networks. There are other examples of widespread tasks also considered in Computational Neuroscience, such as “Perceptual Decision Making” \cite{Britten4745} and “Context-dependent Decision Making” \cite{Mante2013}. Each task has different possible parameterizations. In particular, the task defined in \cite{Mante2013} has recently been used to study the cortex response \cite{Zhang2021}.}

{It is also possible to consider working memory tasks such as “Delay match to sample with two items” \cite{Freedman2006} or “Parametric working memory” \cite{Roitman9475}. For present work, motivated by \cite{DBLP:journals/neco/SussilloB13}, a working memory task, a 3-bit Flip Flop was chosen.}

Once the task is chosen, the requirements must be translated into {an} algorithm that allows us generating the training set. To parameterize the task {considered here}, the following criteria was applied: 

\begin{itemize}

\item The possible states of the Flip Flop are represented in such a way that a positive pulse represents a set and a negative pulse represents a reset. 

\item The state of the output will change corresponding to the input command.

\item A certain delay {in the response} was considered after the falling edge of the input signal.

\end{itemize}

The training data set consists of time series with pulses of fixed duration that represent set and reset signals. Those signals can be activated randomly and are separated by a random time interval. In all time series, a certain noise level has been {added} on the input. Each input elicitate a target output according to the Flip Flop rule: if we have a set signal or positive pulse, the output is in high-state. If we have a reset signal or negative pulse output is in low-state, otherwise, the output remains in the previous state.

The number of inputs in the network corresponds to the number of {memory states} that can be stored. A Flip Flop is a one-bit of memory, {meaning that two states can be stored}. In this case, we have registers formed by three Flip Flops (a 3-bits memory), {which means that we have 8 different memory states}. 

To have the full training data set, it is necessary to generate tensors of size $\texttt{sample\_size}$ with the input time series of length $\texttt{time\_series\_lengh}$ for each of the three inputs and outputs. To do that efficiently, we {used} Numpy arrays \cite{Harris2020}. In the present work, we provide the code to generate a Flip Flop data set. Three random components of the full set $\texttt{x\_train-y\_train}$ are shown in Figure \ref{fig:2}. Input has amplitude noise of $10\%$. The target output, $\texttt{y\_train}$, was simulated with a time delay answer of 20 ms. Each row (and colour) corresponds to one of the inputs and each column to a different sample. Each training sample consists of a Numpy array \cite{Harris2020}. {This is shown in the following code box.}

\begin{lstlisting}
x_train[sample_size,time_series_lengh,3]
y_train[sample_size,time_series_lengh,3]
\end{lstlisting}
\begin{small}
Training data set pairs defined as Numpy arrays.\label{code-3}\\
\end{small}

\begin{center}
\begin{figure}[htb!]
\includegraphics[width=13.5cm]{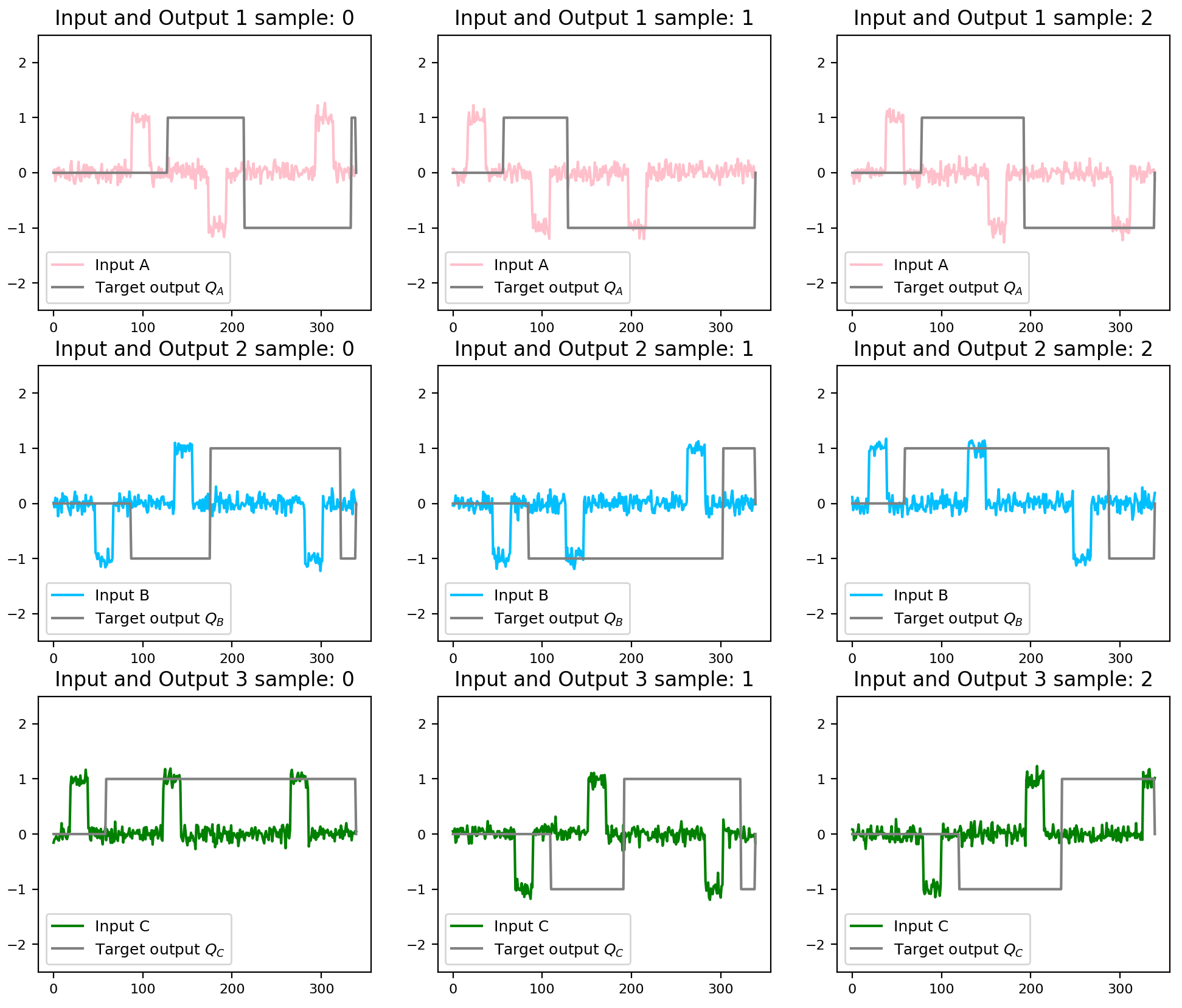}
\caption{{Three random samples of the data set for each input channel. Each row (and colour) corresponds to one of the inputs. Each column corresponds to a different sample. The grey line in each case represents the target output.}}
\label{fig:2}
\end{figure}
\end{center}

\section{Training protocol and parameter selection}\label{traning}

Training methods for neural networks can be unsupervised or supervised. We focused on applying a supervised method.

{Different approaches are available, but those in which a particular type of gradient descent method is applied stand out in the literature. An example is the paradigm of Reservoir Computing, specifically the use of liquid (or echo-state networks) \cite{doi:10.1162/089976602760407955}, where the modifications of the network weights are made in the weights of the output layer, $ \mathbf{W^{out}}$.}

Other outstanding approachs were developed by Sussillo and Abbot. They have developed a method called FORCE that allows to reproduce complex output patterns, including human motion-captured data \cite{susillo_2009}. Modifications to the algorithm have also been applied successfully in various applications \cite{10.1371/journal.pone.0220547, 10.1371/journal.pone.0191527, 10.1371/journal.pcbi.1010590}.

The other method used for estimation of the gradient in RNNs is called Backpropagation Through Time (BPTT), and then an optimization method for minimizing the gradient. Given the recent advances in the implementation of this method with the open-source libraries previously mentioned, this is the method used here. There are also other back propagation-based methods that have been published recently, for example in \cite{Khan2018}, authors propose to use fractional calculus to improve the conventional BPTT.

In this work, supervised learning was used, with standard backpropagation through time implementing an {Adaptive Stochastic Gradient Descent} training method provided by the Keras framework \cite{DBLP:journals/corr/KingmaB14}.

{First, recurrent weights were initialized using a random normal distribution with the orthogonal condition on the matrix. {During} training, noisy square pulse signals were used as the inputs, as the examples shown in Figure \ref{fig:2}}{, and described in Section \ref{task}.}

The appropriate loss function to train the model is the mean square error between the target function and the output of the network. It is defined as:

\begin{center}
\begin{equation}
E(w)=\frac{1}{2}\sum^M_{t=1}\sum^L_{j=1} | \mathbf{Z_j}(t) -\mathbf{Z_j^{target}}(t)|^2,
\label{eq-05b}
\end{equation}
\end{center}

where {$\mathbf{Z_j^{target}}(t)$} is the desired target function and $\mathbf{Z_j}(t)$ is the actual output.

The training {set} consisted of more than $15000$ different {random} samples.
{The previously mentioned training procedures correspond, in terms of the code structure, to the methods for compiling and fitting models. The loss function and the optimizer algorithm are chosen in the compiling step. Different information about the training data set, epochs, and other training characteristics can be specified with the fitting method. An example of implementation is shown in the following code box.}

\begin{lstlisting}

model.compile(loss = "mse", optimizer=ADAM)
model.fit(x_train[50:sample_size,:,:], y_train[50:sample_size,:,:], epochs=epochs, batch_size=128, shuffle=True)

\end{lstlisting}
\begin{small}
Code for the compiling and training steps.\label{code-4}\\
\end{small}

The main parameters of a neural network are the weights of the connections. These parameters are learned during the training stage. On the other hand, hyperparameters are parameters of your neural network that can not be learned via gradient descent or some other training method. These include the learning rate, number of layers, or the number of neurons in a given layer.

Tuning the hyperparameters means the process of choosing the best values of them. Typically this is done by evaluating the performance of the network on a validation set. Then, we have to change the hyperparameters and re-evaluate the model, choosing the values that {gives} the best performance on the validation set. {Another approach for choosing them is to have an informed decision or hypothesis related to the physics or nature of the system under study.}

How do we choose these values? Often there is good standard initialization related to each particular task of interest. An example of the criteria is provided for the Flip Flop {task} in Table \ref{tabla1}.

Another aspect to consider is the regularization of the model. Regularization refers to training our model well enough that it can generalize over data it hasn’t seen before.

To summarize, in the training stage {the main aspects} we have to consider are the size of the network, data set, noise, and regularization terms that are appropriate for the considered task, in our case, the Flip Flop example.

\begin{table}[]
\begin{center}
\begin{tabular}{lclll}
\cline{1-2}
\multicolumn{1}{|c|}{\textbf{Parameter/criteria}} & \multicolumn{1}{c|}{\textbf{Value}}    &  &  &  \\ \cline{1-2}
\multicolumn{1}{|l|}{Units}                       & \multicolumn{1}{c|}{400}               &  &  &  \\ \cline{1-2}
\multicolumn{1}{|l|}{Time step}                   & \multicolumn{1}{c|}{1}              &  &  &  \\ \cline{1-2}
\multicolumn{1}{|l|}{Input Weight}                & \multicolumn{1}{c|}{3 × 400}           &  &  &  \\ \cline{1-2}
\multicolumn{1}{|l|}{Recurrent Weights}           & \multicolumn{1}{c|}{400 × 400}         &  &  &  \\ \cline{1-2}
\multicolumn{1}{|l|}{Output Weight}               & \multicolumn{1}{c|}{400 × 3}           &  &  &  \\ \cline{1-2}
\multicolumn{1}{|l|}{Training algorithm}          & \multicolumn{1}{c|}{BPTT ADAM}         &  &  &  \\ \cline{1-2}
\multicolumn{1}{|l|}{Initialization}              & \multicolumn{1}{c|}{Random Orthogonal} &  &  &  \\ \cline{1-2}
\multicolumn{1}{|l|}{Regularization}              & \multicolumn{1}{c|}{None}              &  &  &  \\ \cline{1-2}
                                              
\end{tabular}

\caption{Model's parameters and criteria for the network's implementation and training.}

\label{tabla1}

\end{center}
\end{table}

\section{Analyzing the results}\label{results}

After training, we obtained a set of RNNs that can perform the tasks {of interest}. Now we {describe in this section} the different aspects to analyze regarding the network's collective behavior. {We selected a method for the model's visualization and a group of tools to extract the relevant information.}

For example, it is possible to visualize the connectivity matrix (recurrent weight matrix), as it is shown on the left side of Figure \ref{fig:3}. The columns represent the output connection of the i-neuron, and the rows are the input connection. {They are also called post-synaptic and pre-synaptic}. The colour bar on the right side represents the intensity of the connections. We have to consider an appropriate scale for the visualization. Even so, it may not be entirely clear how to observe the relevant information, apart from the fact that, after training, most of the weights remain close to zero. {As a first approach for visualization, a plot of the connectivity matrix could be useful, even if the case presented here does not reveal relevant information. It's important not to undervalue it. If the connectivity has some structure farther from a random distribution, it will be observed in the connectivity plot. For example, having null autoconnection terms will be reflected in the color of the diagonal terms of the matrix plot. Another example could be sparsity, which would be reflected in patches over the matrix. Or perhaps, in the case of having excitatory and inhibitory units, it would be easy to visualize the different columns corresponding to the same sign of out connection.}
In case of imposing such constraints on the connections, as Dales' Law \cite{doi:10.1177/003591573502800330}, or any particular constraint they will be visible in this stage, and this representation will be more useful.

{If we don`t obtain relevant information with this first visualization, we know that useful information could be still encoded in the connectivity matrix, but it may not be immediately distinguishable with a connectivity plot. There are different transformations or analyses that can we can perform on the recurrent weight matrix with this aim.} Different Linear Algebra operations are available in the Numpy Library \cite{Harris2020} that are optimized to be used with the array structures. For example, if we perform a decomposition of $ \mathbf {W ^{rec}} $ in their eigenvectors and eigenvalues, we can obtain the eigenvalue distribution as it is shown in the right side of Figure \ref{fig:3}. This analysis can be done using the code in the following code box.

\begin{lstlisting}
from numpy import linalg as LA
eigenvalues, eigenvectors= LA.eig(Matrix)
\end{lstlisting}
\begin{small}
How to use linear algebra library from Numpy for eigenvalue decomposition.\label{code-5}\\
\end{small}

{During training, the matrix associated with the network tend to be non-normal, which results in their eigenvalues to lie closer to the unit circle. This behavior is explained in more detail in papers that study the dynamics of RNNs, where it is shown that the presence of recurrent connections and the attractors in the network's dynamics can cause this accumulation of eigenvalues close to the unit circle \cite{Jarne2022, 10.1371/journal.pcbi.1007655,doi:10.1126/sciadv.aau9403}.}

{Additionally, in these studies, it is typically shown that this accumulation of eigenvalues on the unit circle leads to slowing down the dynamics of the network. They can be linked to the emergence of long-term memories related to the linearization of the system. Therefore, this behaviour can be understood as a necessary condition for the network to effectively store and retrieve information over longer time scales.}

In the case presented here, we can visualize that, except for a small group of eigenvalues that migrated out of the unit circle, the rest remain on it, which is related to the initial orthogonal condition. {The same was replicated throughout all simulations. A set of four examples is shown in Figure \ref{fig:3b}, and the code provided allows us to reproduce more. Eigenvalues outside the unitary circle seem to be related to the behaviour (or modes) observed for the different stimuli at the input as described in {\cite{Jarne2022}}. This is relevant in terms of the dynamics. Additional information related to the connectivity matrix could also be obtained \cite{Jarne2022}.}

\begin{center}
\begin{figure}[htb!]
\hspace{-0.5cm}\includegraphics[width=8cm]{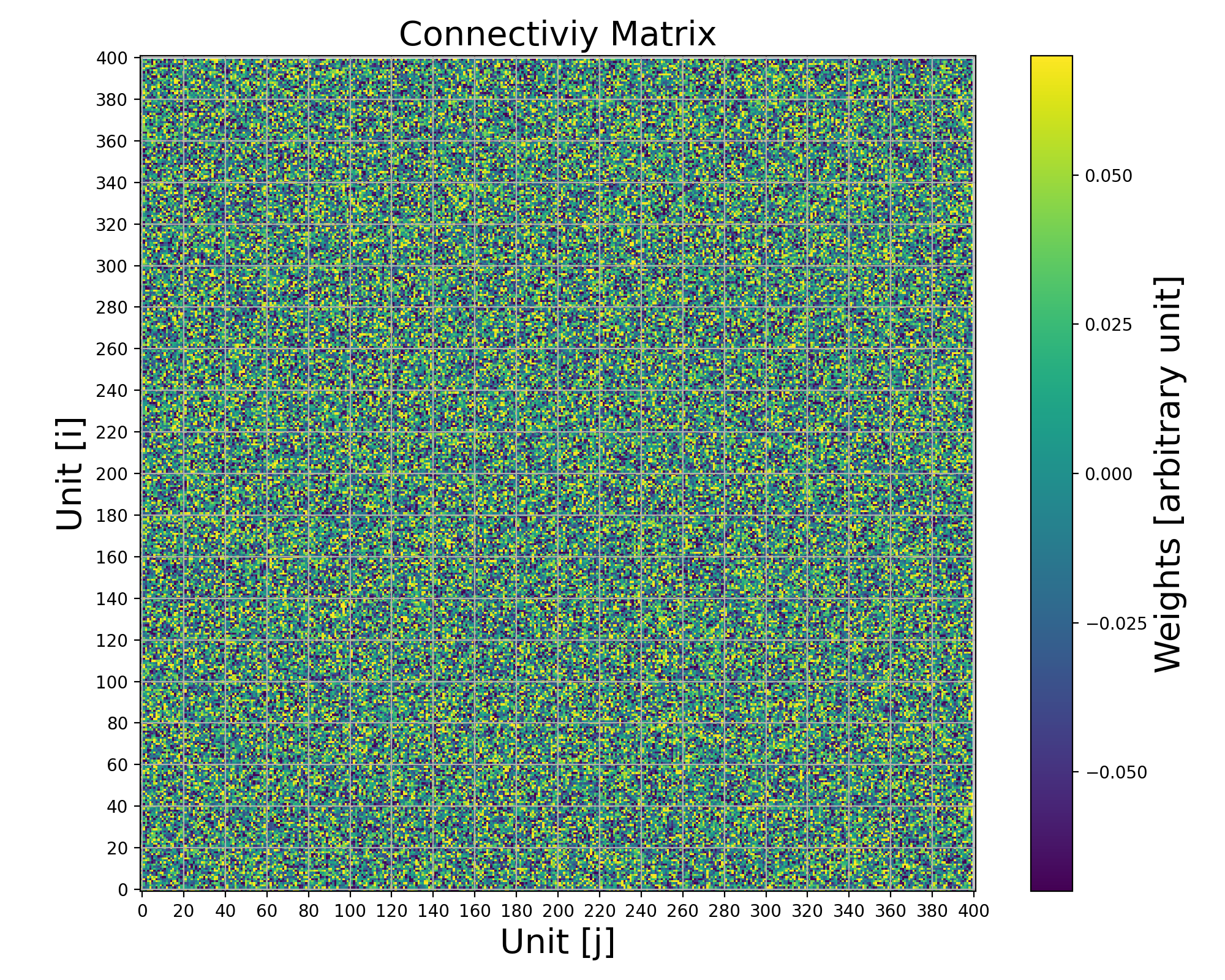}\hspace{-0.2cm}
\includegraphics[width=7.1cm]{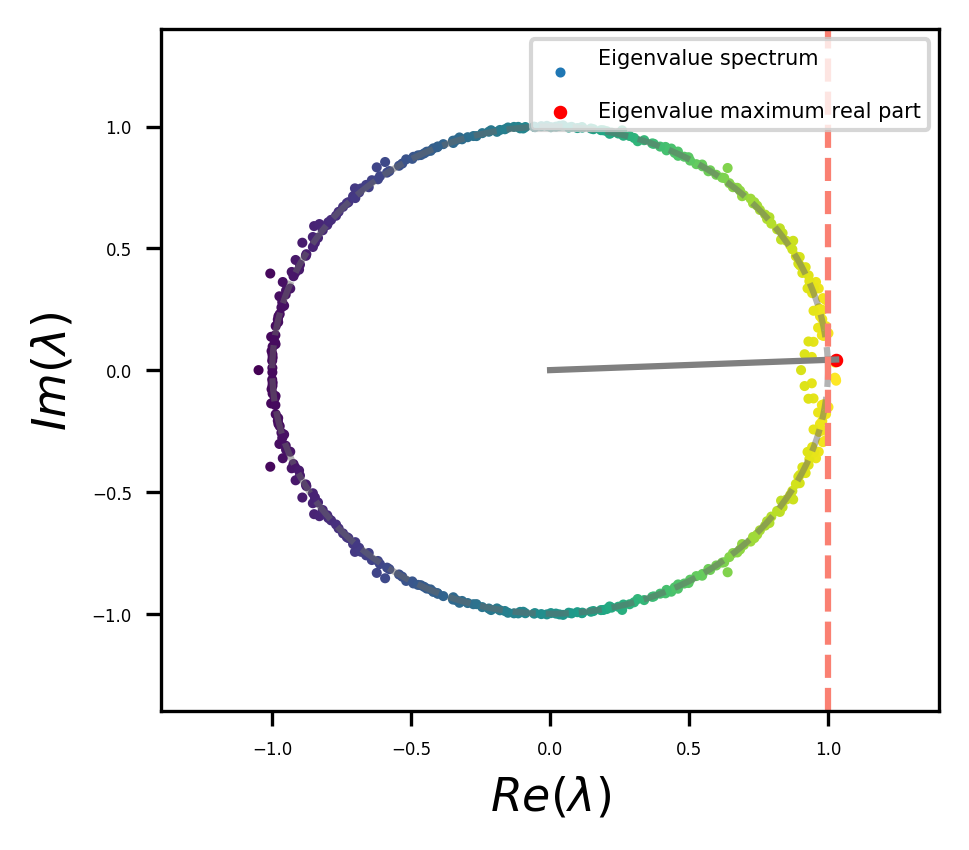}
\caption{Left: Example of the connectivity matrix for a trained RNN. Right: eigenvalue distribution of $ \mathbf {W ^{rec}} $. }
\label{fig:3}    
\end{figure}
\end{center}

\begin{center}
\begin{figure}[htb!]
\includegraphics[width=6.5cm]{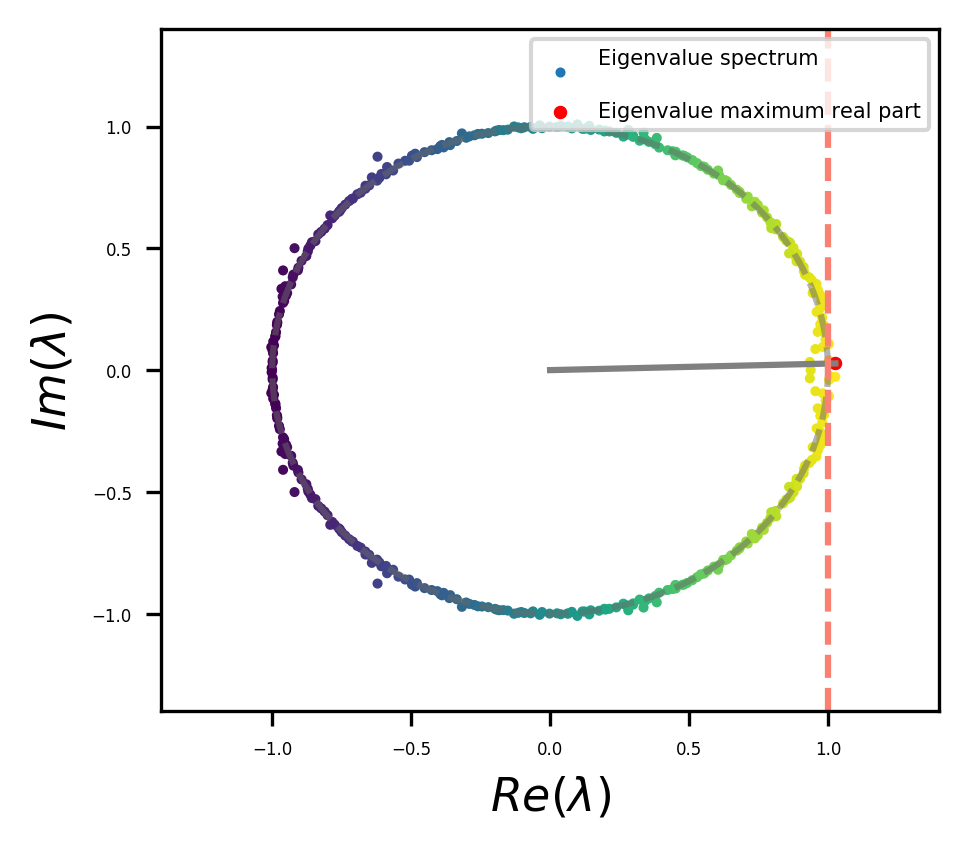}\includegraphics[width=6.5cm]{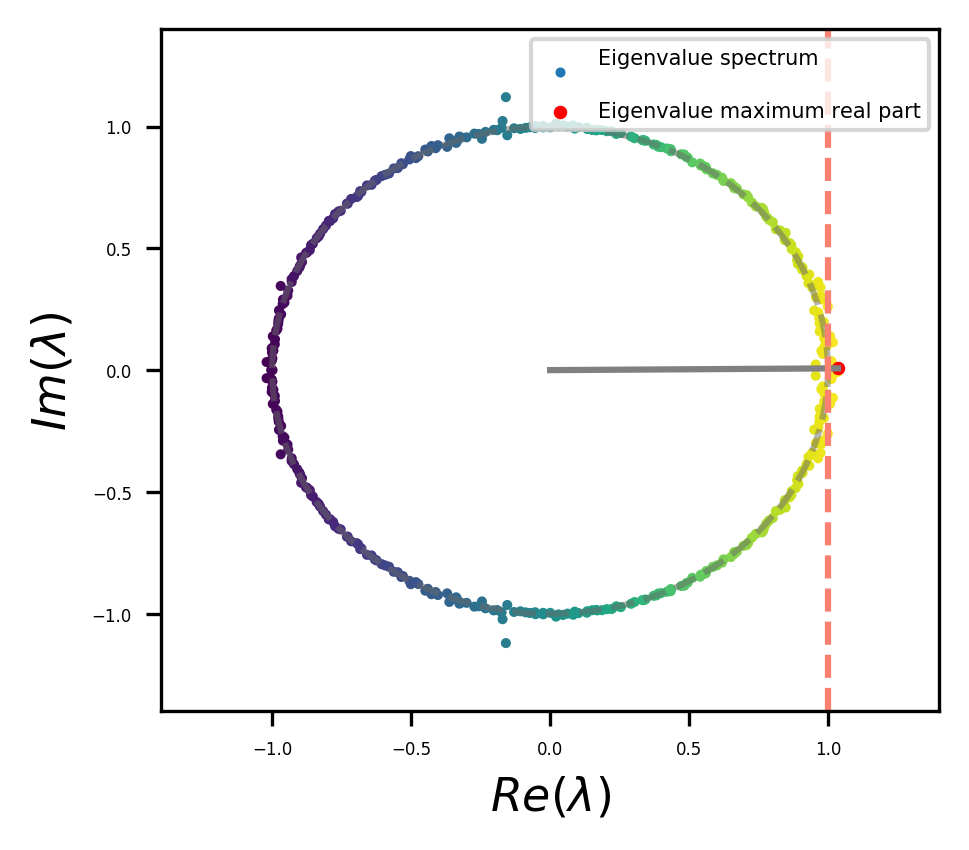}
\includegraphics[width=6.5cm]{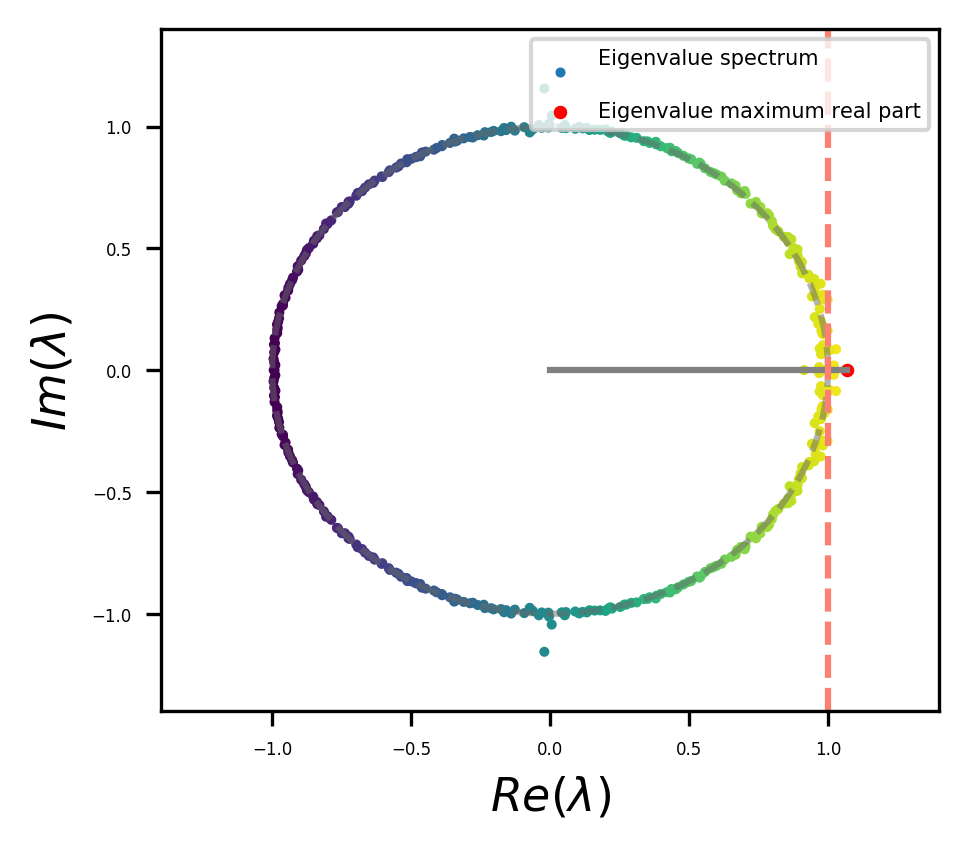}
\includegraphics[width=6.5cm]{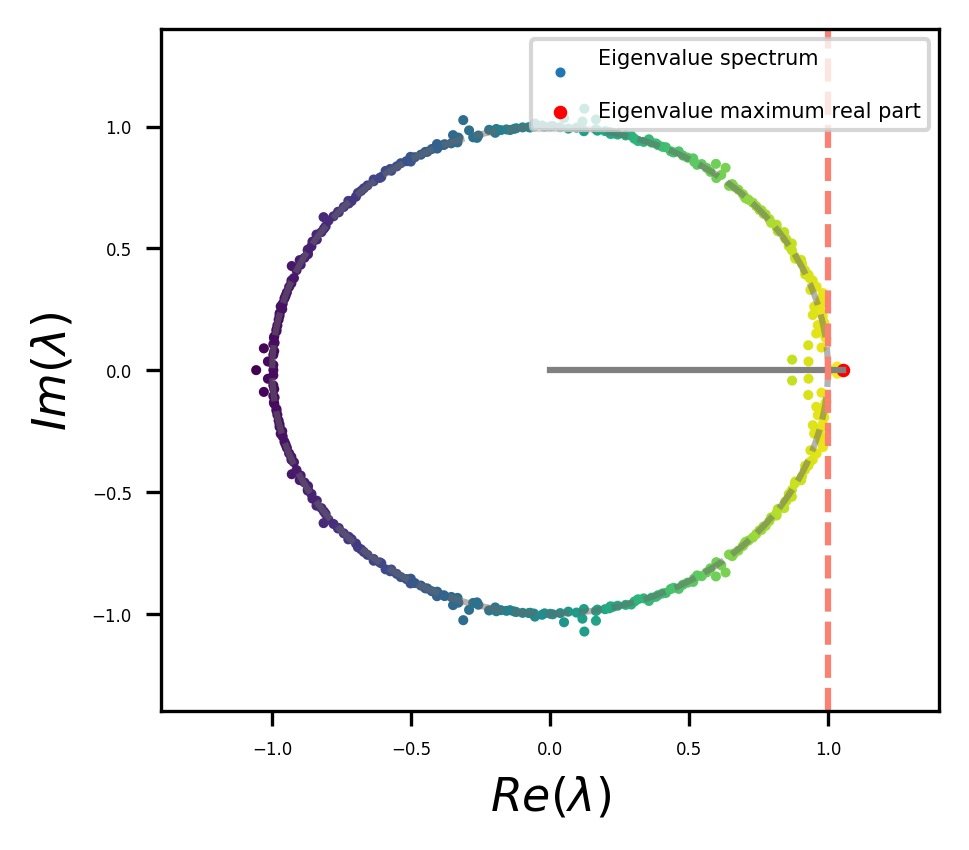}
\caption{Four different examples of eigenvalue distributions of $ \mathbf {W ^{rec}} $ for trained RNNs. }
\label{fig:3b}   
\end{figure}
\end{center}

{Other possible studies that we can perform are related to the response in terms of the activity of the network units when applying the different stimuli.}

Since we have a large number of units, and for each, an activity vector, dimensionally reduction methods are appropriate to analyze such behaviour. {They have} been used widely in different works related to large-scale neural recordings \cite{WILLIAMS20181099, Cunningham2014}.

Scikit-learn \cite{scikit-learn} is a Python open-source library based on Numpy that allows us to perform dimensionality reduction, feature extraction, and normalization, among others. {It has} efficient methods for predictive data analysis. A possible decomposition could be, for example, Principal Component Analysis (PCA) or also Single Value Decomposition (SVD). {The following code box shows how to call the library's functions.}

\begin{lstlisting}
from sklearn.decomposition import PCA
from sklearn.decomposition import TruncatedSVD
\end{lstlisting}
\begin{small}
How to import scikit learn libraries to perform single value decomposition and principal component analysis.\label{code-6}\\
\end{small}

{These tools can be used to extract relevant features of the system. For this work the behaviour, in terms of the activity of the units, was analysed.}

It is well known that the different memory states in a 3-bit memory distribute in the vertex of a cube-like form in the space state \cite{DBLP:journals/neco/SussilloB13}. This was shown when authors explored the hypothesis that fixed points, both stable and unstable, and the linearized dynamics around them, can reveal aspects of how RNNs implement their computations. 

A data set was built for testing and reproducing the behaviour. It generates eight different memory states, as it is shown on the left side of Figure \ref{fig:4}, where each panel shows the input and output of the network. Time series of 600 ms were considered to generate all the different memory states of the 3-bit memory by choosing the correct commutation for the inputs in fixed time intervals. Output responses are shown in red in the Figure.

The testing set is injected into the network (right upper panel of the Figure), and then the activity of the units is analyzed by applying SVD on the activity vector $\mathbf {H}(t)$. The behaviour of the system was represented in the three axes of the greatest variance. The bottom right part of Figure \ref{fig:4} shows the activity in the reduced state space (3-dimensional). Each vertex corresponds to each memory state marked in different colours.

\begin{center}
\begin{figure}[htb!]
\includegraphics[width=14cm]{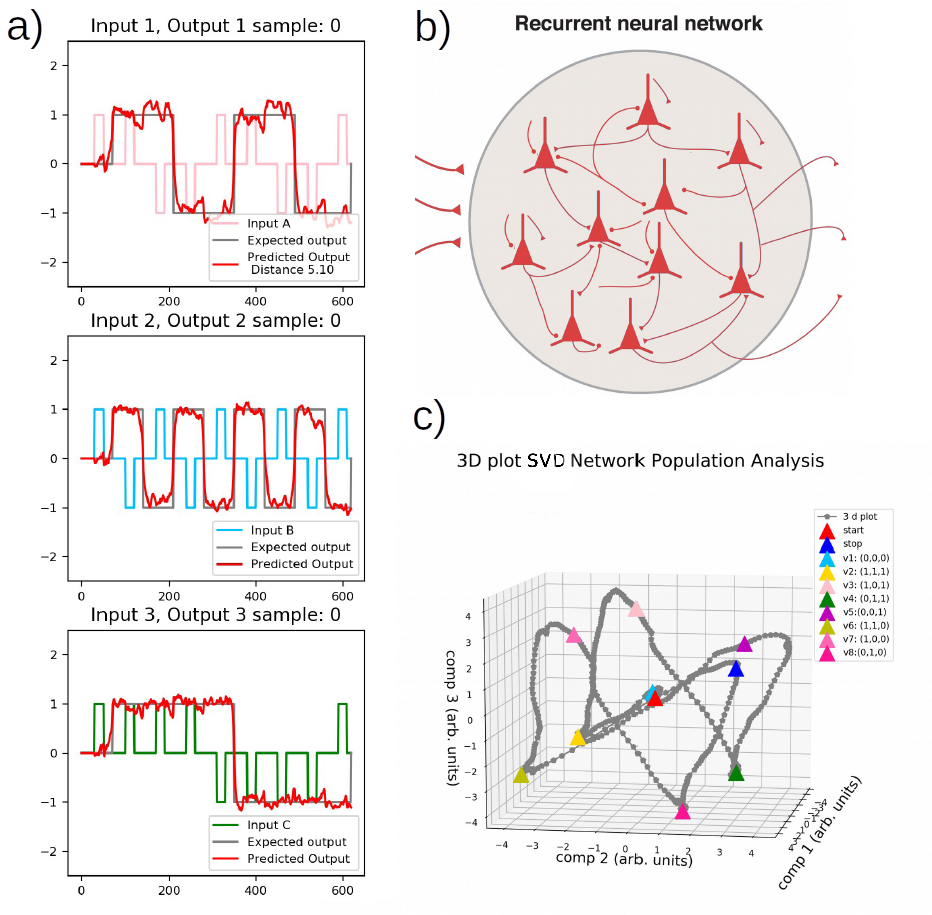}
\caption{ {\textbf{a)} Data set for testing. Each panel corresponds to one input and the predicted output of the Flip Flop, which is shown in red. \textbf{b)} Schema of the network. \textbf{c)} Single Value Decomposition applied activity vector $\mathbf{H}(t)$ of panel b) in the three components of gratest variance. Each colour point corresponds to a different memory state. }}
\label{fig:4}       
\end{figure}
\end{center}

It is well known that different variations of the realizations, in terms of weight distribution and dynamical behaviour, are possible when training networks for the same task {\cite{Jarne2022, 10.1088/2632-072X/abdee3, jarne2019detailed}}. This was exemplified in Figure \ref{fig:3b} and is also shown in Figure \ref{fig:5}, where the four different realizations of the trained networks of Figure \ref{fig:3b} were elicited with the same testing data set and a decomposition SVD analysis, was performed.

{The vertices in this space of the main components are distributed in different positions. A cube-like structure always apears, similarly to what was observed in \cite{DBLP:journals/neco/SussilloB13}, and is rotated in different spatial directions for differen realizations. It is possible to study and classify the behaviour of the obtained systems by comparing the the network obtained. This cube-like structure is characteristic of this task parametrization, and it appears even when here we used a different training method and network parameters compared with previus studies \cite{DBLP:journals/neco/SussilloB13}.}

{Additional analysis could be considered depending on the aspects of interest to be studied. Here a minimal analysis was proposed.} We described in detail the steps, visualization tool, criteria, and implementation. The code for training and analysis is provided. Also it can be used as an open framework to parametrize different tasks or additional studies. {In this way, we can compare the different realizations for the Flip Flop task.}

{}

\begin{center}
\begin{figure}[htb!]

\hspace{-1cm}\includegraphics[width=7.81cm]{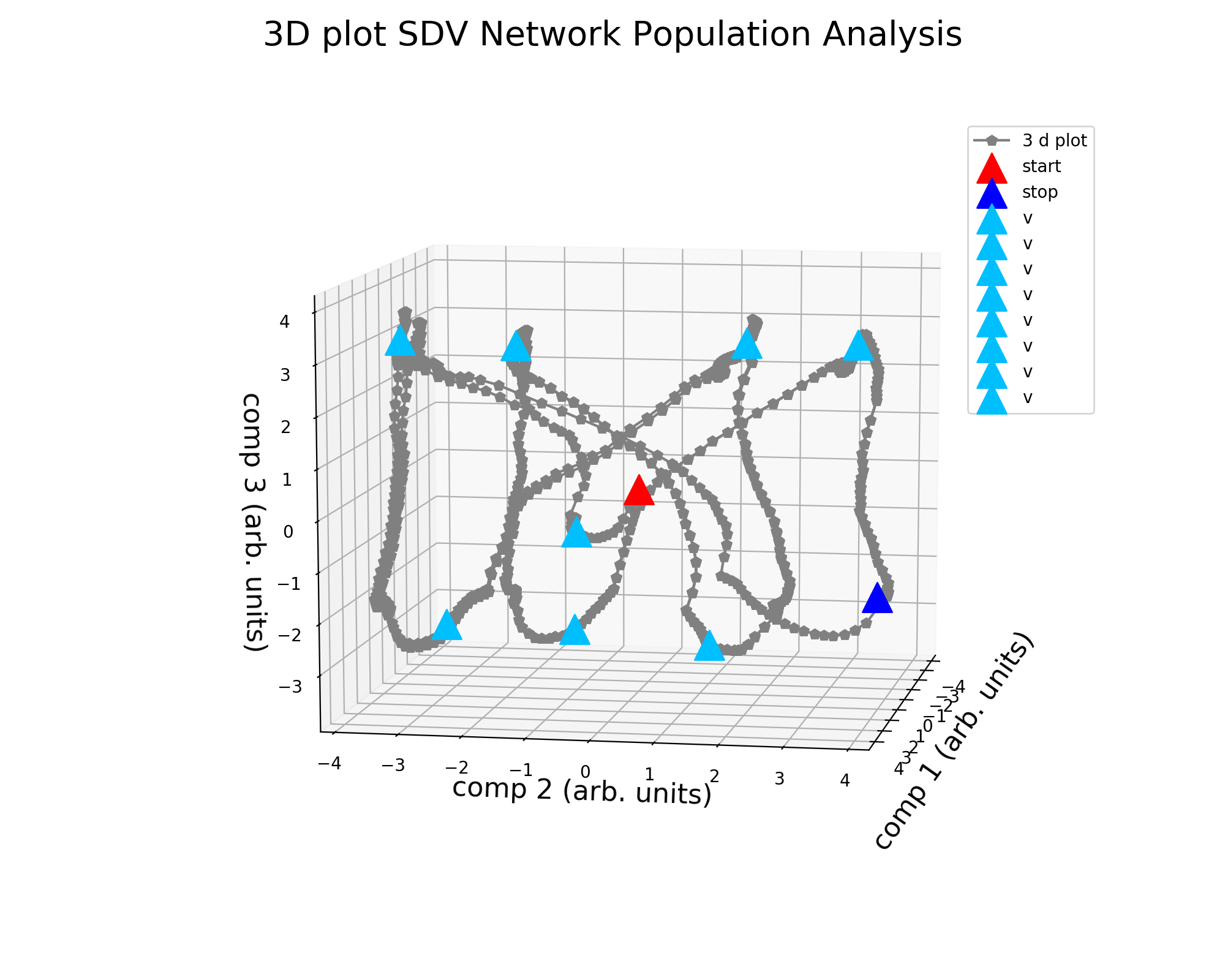}
\hspace{-1cm}\includegraphics[width=7.81cm]{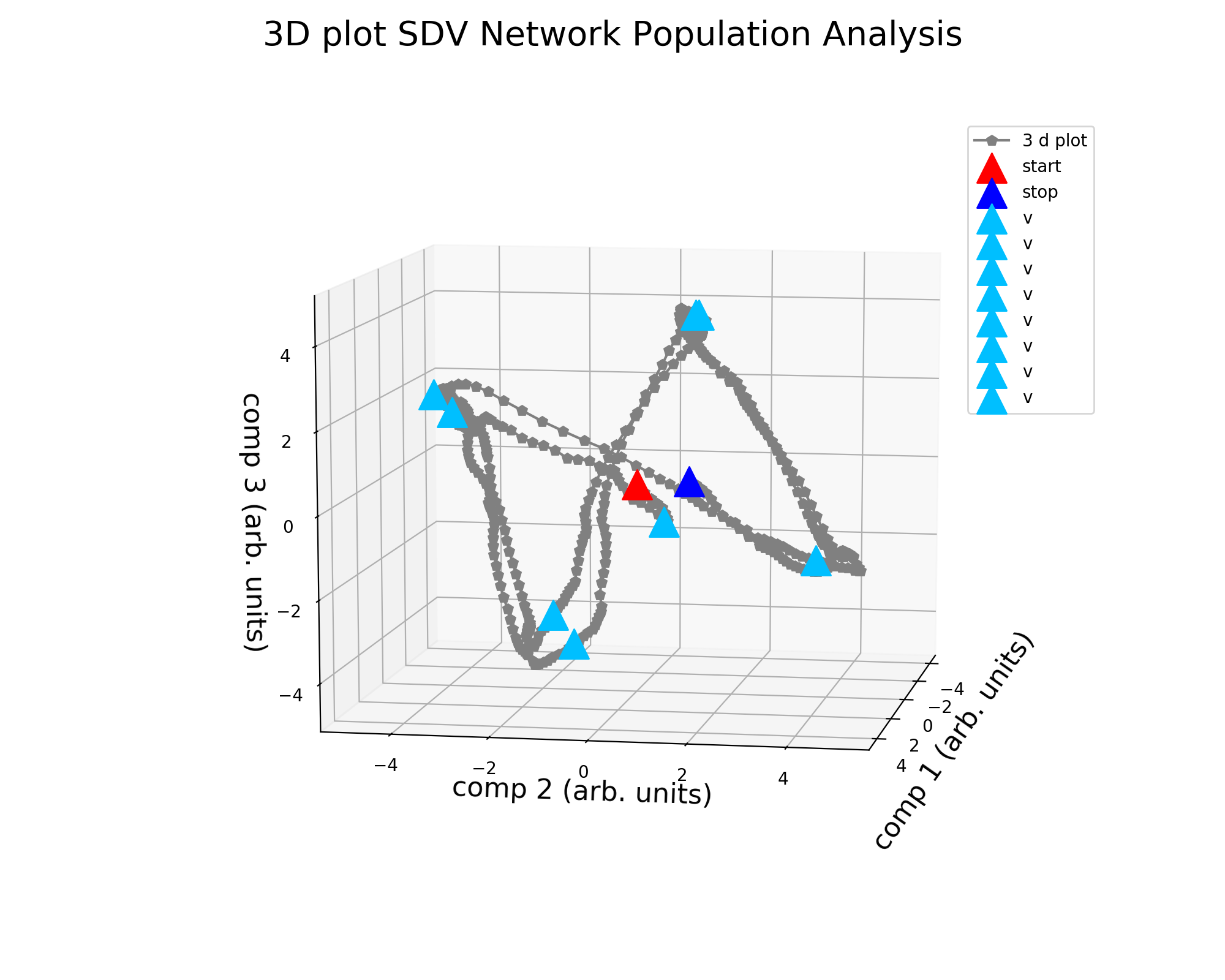}\\

\hspace{-1cm}\includegraphics[width=7.81cm]{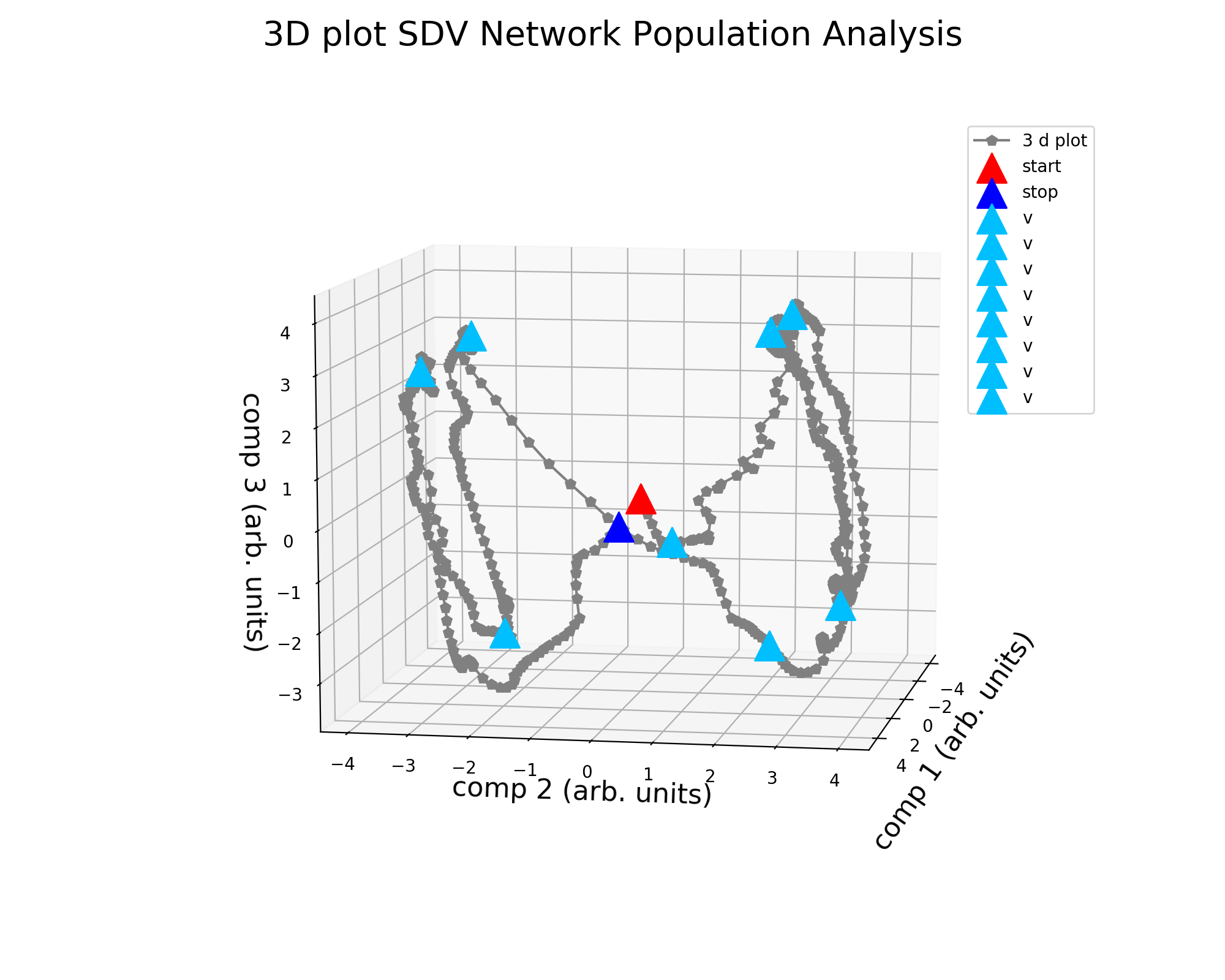}
\hspace{-1cm}\includegraphics[width=7.81cm]{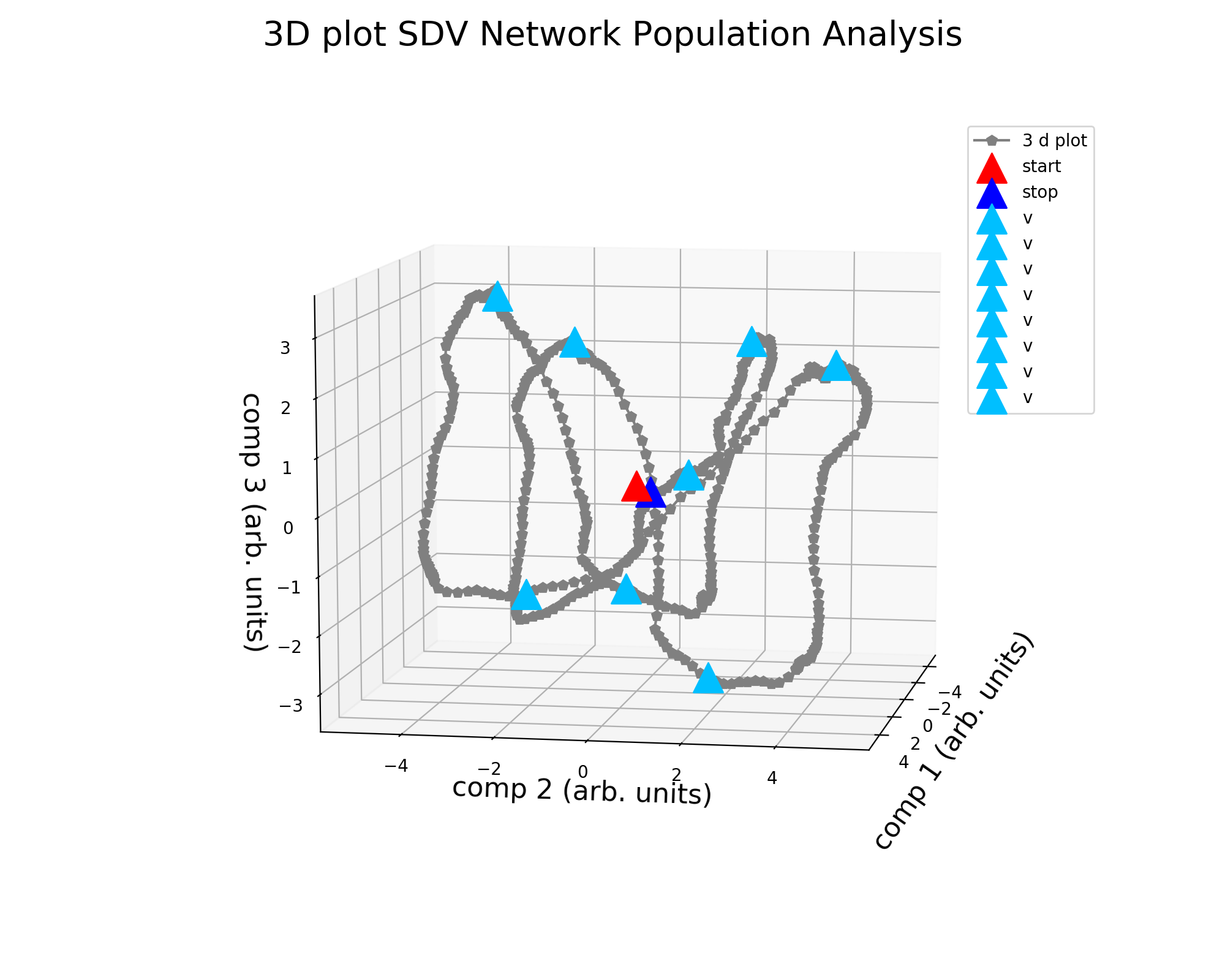}

\caption{Activity of the four different realizations of RNNs trained for the same 3-bit Flip Flop task coresponding to Figure \ref{fig:3b}.}
\label{fig:5}  
\end{figure}
\end{center}

\section{Conclusions}\label{conclu}

In this work, all steps to build and analyze an RNN have been presented for a sample task. We started from the model description in terms of the equations, discretization, and code implementation. {We discussed different options that are available for code implementation depending on the considered model and scientific questions.} Then, we described the task parametrization and network training protocol. We also presented a set of tools to analyze the results using open-source scientific libraries making use of the different visualization tools that allow extracting relevant features.

We used the Flip Flop task as an example, but other relevant tasks could be considered, { as mentioned in Section \ref{task}. For example, “Perceptual Decision Making” \cite{Britten4745}, “Context-dependent Decision Making” \cite{Mante2013, Zhang2021},  working memory tasks such as “Delay match to sample with two items” \cite{Freedman2006} or “Parametric working memory” \cite{Roitman9475}.}  In this work, motivated by \cite{DBLP:journals/neco/SussilloB13}, a working memory task such as a 3-bit Flip Flop, { was chosen to show the full process: from the differential equations of the RNN model, discretization, through the parameterization of the task and the methods of analysis for the activity of the network against the different stimuli on the network.}

The use of the open-source scientific tools that are designed and maintained for large communities, such as the tools used here, allows enhancing research. This is why we are currently using tools that are more transparent in terms of code and documentation, because they are open to being modified and improved by thousands of users.

\section*{Conflict of interest}

The author declare that she has no conflict of interest.

\section*{Code availability}
Code is provided in an open repository after paper publication and with a MIT License.
\url{https://github.com/katejarne/3-bit-FF-tutorial}

\section*{Acknowledgments}

The present work was supported by CONICET and UNQ. Author acknowledges support from PICT 2020-01413. 

\Urlmuskip=0mu plus 1mu\relax

\bibliographystyle{elsarticle-num} 
\bibliography{mybibfile.bib}

\end{document}